\documentclass[runningheads]{llncs}
\usepackage[T1]{fontenc}
\usepackage{rotating}
\usepackage{adjustbox}
\usepackage{graphicx}
\usepackage{amssymb}
\usepackage{algorithm,multirow}
\usepackage{tikz}
\usepackage{ctable}
\usepackage{comment}
\usepackage{makecell}

\usepackage[noend]{algpseudocode}

\usepackage{subfig}
\begin{document}
\title{MTS2Graph: Interpretable Multivariate Time Series Classification with Temporal Evolving Graphs}
\titlerunning{MTS2Graph: Interpretable Multivariate Time Series Classification}
%\author{Raneen Younis\inst{1}} \and
%Second Author\inst{2,3}\orcidID{1111-2222-3333-4444} \and
%Third Author\inst{3}\orcidID{2222--3333-4444-5555}}
%
\authorrunning{Younis et al.}
%\institute{Princeton University, Princeton NJ 08544, USA \and
%Springer Heidelberg, Tiergartenstr. 17, 69121 Heidelberg, Germany

\author{Raneen Younis\inst{1} \and
Abdul Hakmeh\inst{2} \and
Zahra Ahmadi\inst{1}}
\authorrunning{Younis et al.}
\institute{L3S Research Center \email{\{younis,ahmadi\}@l3s.de} \and
University of Hildesheim
\email{hakmeh@bwl.uni-hildesheim.de}}

\maketitle              % typeset the header of the contribution
\begin{abstract}
Conventional time series classification approaches based on bags of patterns or shapelets face significant challenges in dealing with a vast amount of feature candidates from high-dimensional multivariate data. In contrast, deep neural networks can learn low-dimensional features efficiently, and in particular, Convolutional Neural Networks (CNN) have shown promising results in classifying Multivariate Time Series (MTS) data. A key factor in the success of deep neural networks is this astonishing expressive power. However, this power comes at the cost of complex, black-boxed models, conflicting with the goals of building reliable and human-understandable models. An essential criterion in understanding such predictive deep models involves quantifying the contribution of time-varying input variables to the classification. Hence, in this work, we introduce a new framework for interpreting multivariate time series data by extracting and clustering the input representative patterns that highly activate CNN neurons. This way, we identify each signal’s role and dependencies, considering all possible combinations of signals in the MTS input. Then, we construct a graph that captures the temporal relationship between the extracted patterns for each layer. An effective graph merging strategy finds the connection of each node to the previous layer’s nodes. Finally, a graph embedding algorithm generates new representations of the created interpretable time-series features. To evaluate the performance of our proposed framework, we run extensive experiments on eight datasets of UCR/UEA  archive, along with HAR and PAM datasets. The experiments indicate the benefit of our time-aware graph-based representation in MTS classification while enriching them with more interpretability.
\end{abstract}
\section{Introduction}
Development of sensor usage in various applications such as medical care, activity recognition, and weather forecasting has raised our access to time series data~\cite{sharabiani2017efficient,ullah2013applications}. Collecting and processing such a large volume of temporal (or spatio-temporal) data from these applications has led to the development of various data mining methods. In many of these applications, the collected data contains more than one sensor value at a time, and the sensor values may depend on each other. This form of data is called multivariate time series (MTS) and has several challenges due to the necessity of extracting the hidden and unknown temporal relationship between data points and signals. There is a rich literature on multivariate time series classification methods, from a simple nearest neighbor~\cite{orsenigo2010combining}, %which has even shown promising results in multivariate time series classification~\cite{orsenigo2010combining}, 
to dynamic time warping, which various studies have shown its best classification results compared to the distance-based algorithms~\cite{seto2015multivariate}. 
Feature-based methods are a classification algorithm family that utilizes feature extraction to improve prediction performance by identifying distinctive features from a long time series sample. These methods divide time series data into equal-sized segments and extract statistical properties~\cite{nanopoulos2001feature} or apply transformation functions such as wavelet transformations~\cite{d2014wavelet}. Feature extraction can also be accomplished via time-series shapelets~\cite{ye2009time} or bag of features~\cite{baydogan2013bag}, resulting in an extended feature space that may reduce classification accuracy~\cite{schafer2017multivariate,zhang2020tapnet}. Multivariate time series feature extraction is complex, and the preprocessing step adds a computational burden. Therefore, distance-based algorithms are preferred over feature-based methods, as they perform better~\cite{zheng2014time}.

%Feature-based methods are another classification algorithm family where a feature extraction step plays a significant role in enhancing the prediction performance by ideally finding distinctive features from a long time series sample. In their basic form, they divide time series data into equal-sized segments and apply a feature extraction method to each segment to extract the statistical properties of a segment~\cite{nanopoulos2001feature}, or apply a transformation function such as wavelet transformations~\cite{d2014wavelet}. Feature extraction can be accomplished via time-series shapelets~\cite{ye2009time}, or bag of features~\cite{baydogan2013bag} where the data is converted to a set of subsequences or patterns to be used as feature candidates for the classification task. The generated features result in an extended feature space, which can be huge and may reduce the multivariate time series classification accuracy~\cite{schafer2017multivariate,zhang2020tapnet}. On the other hand, multivariate time series feature extraction is a complicated task because of the complexity of capturing the inherent features of time series. Moreover, the preprocessing step for feature extraction from long time series adds a computational burden to the classification. Therefore, the distance-based algorithms are preferred in this case and perform better than the feature-based methods~\cite{zheng2014time}. 

\begin{figure}[t]
\centering
\includegraphics[scale=0.60]{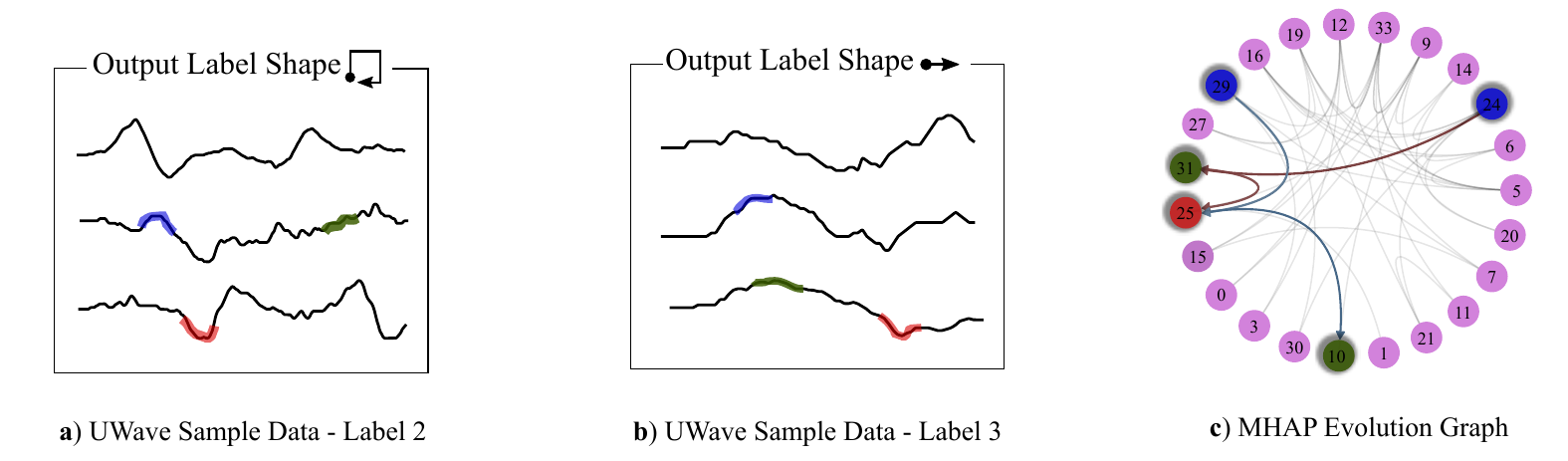}
%\includesvg{images/MotivationExample.svg}
\caption{a) and b) two instances of the UWave dataset with their respective highly activated subsequences from the first CNN layer, c) the \textit{Multivariate Highly Activated Period (MHAP) evolution graph} of the network's first layer that is built based on the timely occurrence order of the extracted MHAPs.} \label{motivationImage}
%\caption{a) An instance of the three-variate UWave dataset with the output class $2$ and its respective highly activated subsequences from the first CNN layer highlighted in blue, green, and red, b) another instance of the UWave dataset with the output class $3$ and its highly activated subsequences, c) the \textit{Multivariate Highly Activated Period (MHAP) evolution graph} of the network's first layer that is built based on the timely occurrence order of the extracted MHAPs. The colored nodes represent the respective MHAPs of the two instances in a and b with their corresponding edges.} \label{motivationImage}
\end{figure}

Instead of employing a separate feature extraction module, one may consider deep neural networks. Deep learning has shown promising results in various applications in recent years \cite{lecun2015deep}, including time-series data \cite{ismail2019deep}. Due to their end-to-end classification nature, deep models do not rely on heavy feature engineering processes. Different deep architectures were used for multivariate time series classification, such as Multivariate Long Short Term Memory Fully
Convolutional Network (MLSTM-FCN)~\cite{karim2019multivariate}, Convolutional Neural Networks (CNN)~\cite{zheng2016exploiting}, and Recurrent Neural Networks (RNN)~\cite{che2018recurrent}. These models can learn low-dimensional feature representatives using recurrent or convolutional networks, reducing the problem of large feature space arising in traditional classification methods. Furthermore, deep learning methods depend less on domain knowledge and can be applied to different time series applications. Despite all the advantages of deep neural networks, one main disadvantage of these models is their black-box nature, where their inner functioning and decisions are ambiguous to users and even their designers. The lack of interpretability of the deep learning models is problematic in some applications such as law, finance, or medicine. In such applications, the explainability of the model's decisions is essential in building trustworthy and reliable models and comprehending the model's predictions from a human perspective~\cite{hsieh2021explainable}. Several studies have introduced model interpretability, focusing mainly on image data \cite{zhou2016learning,montavon2017explaining,ribeiro2016should}. However, interpretability for multivariate time series data is rather unexplored while it has its own challenges considering the specific nature of multivariate time series data that encodes the patterns from different variables over time.

In this paper, we introduce a novel framework, MTS2Graph, for interpreting decisions generated by deep neural network models for multivariate time series data. For that, we focus on convolutional neural networks and extract the representative patterns from the input MTS data, which highly activate neurons in the CNN-based network. We call these patterns Multivariate Highly Activated Period (MHAP). These MHAPs are then grouped using K-shape clustering \cite{paparrizos2015k} to generalize the extracted patterns better. Then, we create a graph representing the extracted MHAPs in the temporal order, rather than solely considering them at their occurrence time. Each graph node represents an MHAP, a cluster median, and an edge represents the chronological occurrence of an MHAP after another.  
Fig.~\ref{motivationImage} shows two samples from the UWave dataset tested in our framework. The first instance (Fig.~\ref{motivationImage}a) contains three representative patterns (nodes $29$, $25$ and $10$) learned from the first layer of the network and the second instance (Fig.~\ref{motivationImage}b) consists of representative MHAP nodes $24$, $31$, and $25$. The graph in Fig.~\ref{motivationImage}c represents the relationship between these nodes based on their temporal occurrence order. We observe that node $25$ occurs in both samples. However, the temporal order is different in the two classes, and this is where our graph representation becomes helpful by identifying the representative patterns and their chronological order that lead the classifier to a specific decision. The main contributions of the MTS2Graph framework are as follows:
\begin{itemize}
    \item We extract and visualize the representative patterns learned by the CNN channels and consider the entire multivariate time series signals and their correlation.
    \item We build an MHAP evolution graph that keeps the temporal relation between the extracted representative patterns for each network layer and then merge these graphs into one. 
    \item A graph embedding algorithm is applied to the graph, and a new classifier is built for the new feature representations.
    \item We evaluate the predictive performance of our approach via extensive experiments on eight datasets from the well-known UCR/UEA archive benchmark~\cite{dau2019ucr,baydogan2019multivariate},
    %\cite{baydogan2019multivariate},
    along with HAR \cite{anguita2013public} and PAM \cite{reiss2012introducing} datasets.
\end{itemize}
The remainder of the paper is organized as follows: In Section~\ref{relatedwork}, we discuss the related work in the context of time series classification and the interpretable machine learning methods. Section~\ref{method} describes our proposed MTS2Graph framework. In Section \ref{results}, we evaluate the framework's performance on different datasets. Finally, Section~\ref{conclusion} concludes the paper and indicates directions for future work.

\section{Related Work}\label{relatedwork}
\subsection{Multivariate Time Series Classification}
Multivariate time series classification has become a prevalent task due to data available from various applications. Existing approaches can be categorized into three main groups: distance-based methods, feature-based methods, and deep learning-based methods. The first category, distance-based methods, classifies time series data based on a similarity measure like nearest neighbor (NN) or dynamic time warping (DTW). DTW and NN provide a robust baseline for multivariate time series classification \cite{bagnall2017great}. 
Another category of classifiers uses feature-based methods to extract representative features from time-series data \cite{karlsson2016generalized,li2021shapenet}. For example, the Ultra-Fast Shapelets (UFS) method extracts representative shapelets from data to discriminate between classes efficiently~\cite{wistuba2015ultra}. $GMSM_{red}$ is another shapelet-based approach that learns multivariate shapelets from time series input data \cite{medico2021learning}.
Further, Schaefer and Leser present WEASEL-MUSE~\cite{schafer2017multivariate}, the state-of-the-art shapelet-based method which uses a bag of symbolic Fourier approximation. Feature-based methods assume that user knowledge about the time series data exists and require handcrafted feature engineering, which can be complicated for long multivariate time series. 
The success of the deep learning method in different applications motivated researchers to adapt them to time series data \cite{ismail2019deep,karim2019multivariate,zhang2020tapnet,bai2021correlative}. For instance, CNN-based models such as Multi-Channel Deep Convolutional Neural Network (MC-DCNN) \cite{zheng2016exploiting}, Residual Network (ResNet) \cite{he2016deep}, and Fully Convolutional Networks (FCN) \cite{long2015fully} are used for time series classification tasks, and they show remarkable results \cite{ismail2019deep}. Despite their impressive results in time series classification tasks, interpreting deep learning outputs remains a challenge.
%Another shapelet-based approach, called Generalized Random Shapelet Forests (gRSF), constructs a decision tree based on randomly selected shapelets~\cite{karlsson2016generalized}. 
\subsection{Interpretable Models}
In recent years, study on the interpretability of complex learning methods, such as deep learning models, has become widespread. Post-hoc methods are a famous family of interpretable methods which require a trained and fixed-targeted model for interpretation. Some examples include: Local Interpretable Model-agnostic Explanations (LIME) \cite{ribeiro2016should}; gradient-based approaches \cite{selvaraju2017grad,shrikumar2017learning}, %which estimate the effect of a change in input features on model predictions; 
and Class Activation Map (CAM) \cite{zhou2016learning}.
%which visualizes the most contributing parts of the input data to the classification decision.
These post-hoc interpretation methods have remarkably succeeded in computer vision and natural language processing. As they visualize the impact of a particular pattern in the input data on the output decision, humans can easily understand these parts of the input data.
Wang \textit{et al.} adapted the CAM method to time series data to visualize the region of raw input data that activates the network's neurons for a given label  \cite{wang2017time}. In another work on univariate time series data, Clustered Pattern of Highly Activated Period (CPHAP) \cite{cho2020interpretation,cho2021interpreting} extracts those representative patterns from the input data that highly activate neurons in a CNN channel. However, all these methods can only handle univariate time series data.

Interpretability by design is another interpretable method that focuses on creating an interpretable and accurate classifier. \cite{cheng2020time2graph} present the Time2Graph framework for univariate time series data, where a dynamic shapelet extracts features from the data and creates a shapelet graph that captures the evolution of the shapelets over time. However, Time2Graph requires handcraft feature engineering. Another shapelet-based classifier for univariate time series data introduced by\cite{le2023learning}. Feremans \textit{et al.} \cite{feremans2022petsc} introduced a dictionary-based method for time series classification and applied their feature extraction way for each dimension independently.
A Dual-stage Attention-based Recurrent Neural Network method (DA-RNN) that considers the long-term temporal dependencies and selects the relevant series deriving the model predictions via an input-attention mechanism presented by \cite{qin2017dual}. \cite{xu2018raim} introduce a Recurrent Attentive and Intensive Model (RAIM) to analyze the continuous monitoring of discrete clinical events. \cite{guo2019exploring} explore the structure of recurrent LSTM neural networks to learn variable-wise hidden states to capture the role of different variables in predicting multivariable time series models. 
Gated Transformer Networks (GTN)~\cite{liu2021gated} were introduced for multivariate time series data by adding gating to the Transformer networks. The authors investigated the interpretability of the model by exploring the GTN attention map on time series data. Unlike the previous studies, we augment a CNN-based architecture by extracting the most relevant time intervals to the prediction and creating a graph based on the evolution of the extracted critical time-series sequences over time. This way, we benefit from no requirement for handcrafted feature engineering or domain knowledge, consider the correlation among variables in the multivariate time series, and keep track of the chronological order of extracted features.

Recently, \cite{hsieh2021explainable} presented an interpretable convolutional-based classifier for multivariate time series data that adds an attention mechanism to the extracted features to identify the signals and the time intervals that select the classifier output. 
%\footnote{Although the method is a has similarities to our framework,  Unfortunately, the code is not public for further experiments and results reproducibility.}
The main advantage of Hsieh's method compared to the previous methods is that it identifies both the relevant subset of time series variables and the time intervals in which the variables contribute to the distinction between classes. However, it does not indicate how the identified subset of time intervals evolves over time nor considers the correlation between multivariate time series signals. This method is a suitable comparison method to our framework due to its focus on the interpretability of CNN models for multivariate time series data; however, the authors do not publicly provide the code for further experiments and reproducibility.

\section{MTS2Graph Framework}\label{method}
This section presents the path to building an interpretable multivariate time series classifier based on the outputs of a CNN classifier and with the benefit of a time-aware evolution graph. First, MTS2Graph extracts those time series periods from the input multivariate time series (MTS) that highly activate neurons in a specific CNN model; these periods are considered the selected representatives and the most important features to the model (Sec.~\ref{MHAP}). Then, a graph is constructed with nodes indicating these time series representatives and edges capturing their evolution in a time series sample (Sec.~\ref{Graph}). Later, an embedding algorithm transforms the graph into a latent feature space where new representations are constructed based on the generated latent feature space vector, and classification is performed on this embedded space (Sec.~\ref{learning}).

\subsection{Multivariate Highly Activated Period Extraction}\label{MHAP}
The decision of a trained deep learning network model is usually influenced by its highly activated neurons. These neurons are activated based on the features captured by the training data. 
Cho \textit{et al.} \cite{cho2020interpretation} proposed a method to extract the highly activated periods from univariate time series data, while \cite{younis2022multivariate} proposed a similar method for multivariate time series data.
The unique features of multivariate time series data, i.e., unknown dependencies among multiple channels (dimensions) and through time, pose challenges.
Therefore, similar to \cite{younis2022multivariate}, we extract the patterns from all signal combinations to capture and maintain the existing correlations in the multivariate time series data. Thus, we extract the learned features through a neural network from each signal and signal combination. MTS2Graph proposes a new interpretability by design classifier based on the representative patterns learned by the CNN channel. However, in \cite{younis2022multivariate}, the author proposed a method that visualizes the patterns learned by a CNN network, and no changes were made to the classification method. 
%However, MTS2Graph proposes a new interpretability by design classifier based on the representative patterns learned by the CNN channel.
%Table~\ref{table4:term} shows the definitions of the frequent terminologies used in this paper.
%\input{images/table_term}

The multivariate highly activated period (MHAP) extraction process is divided into three steps: First, a convolutional neural network is trained on the multivariate input data. Besides that, an input set is created containing all possible signal combinations of a multivariate data. This input set is then passed to the second step, which extracts  MHAPs from a trained model. In the third step, the extracted MHAPs are clustered to combine similar subsequences, and a cluster median is associated with each cluster. In the following, we will explain these steps in more detail:

\subsubsection{Trained Model and Extraction of the input set:}
We train a 1D-CNN model that contains more than one convolutional layer. The input set contains either one or a combination of several multivariate time series signals concatenated in a 1D vector since the multivariate data may be partially dependent or mutually independent. A signal is extracted from the original data by leaving its values untouched and setting all other signals to zero. Likewise, the signal combinations are created by leaving the original required multivariate signals untouched and setting all other signals to zero.   
\subsubsection{Extraction of the Highly Active Periods:}
To extract the input subsequence that highly activates neurons in the CNN model, we first identify the highly active neurons. A CNN model consists of layers, $l$, that have different channels or filters, $c$, and we analyze these neurons for each channel of each hidden layer. Hence, we define a threshold $T_{l,c}$ for the channel $c$ and layer $l$ that satisfies $P(activations_{l,c} >T_{l,c}) = 0.05$. A neuron $n$ with an activation value $A(n)$ greater than this threshold is considered activated and is then added to the list of highly activated neurons (HAN):
\begin{equation}
HAN = [
\forall n \in [1, N_{l,c}] ; \; \emph{return} \; \textbf{n} \; where \; A(n) \geq T_{l,c}],
\end{equation}
where $N_{l,c}$ represents all the neurons at channel $c$ in the hidden layer $l$. 

Based on the extracted HAN, the corresponding MHAPs are extracted from the input data:
\begin{equation}
MHAP = [Signal\; Receptive \;Field(SRF) \;of \;i\; where \;i\in {\it MHAP}_{l,c} ].
\end{equation}

\subsubsection{Clustering Data:}
Each CNN layer usually provides similar MHAP patterns that can be grouped together. For this purpose, we apply the K-shape clustering algorithm \cite{paparrizos2015k} to each layer's MHAP data and preserve the cluster median (CM) as a representative pattern for a cluster:
\begin{equation}
CM_{MHAP} = [p_{i,j} \; where \; p_{i,j} \in Median\{K_{shape}(MHAP)\}], 
\end{equation}
where $p_{i,j}$ is the highly activated sequence of sequence $j$ which represents {\it MHAP}s of each input channel $i$.

Among various clustering methods in the literature, such as hierarchical clustering, k-mean, K-shape, and many more, we employ the K-shape method because of its remarkable properties for clustering time-series data, such as noise invariance and dealing with phase shift in data.

%\input{images/Algorithem1}
%Algorithm~\ref{alg:MHAPsudocode} explains the entire process of extracting the MHAP from the input data. The algorithm takes the training data as input and returns the extracted MHAP and the cluster medians as output. 
%First, a CNN-based model is trained using the training data (line $1$). A loop is executed for each training sample (lines $2$-$9$) and extracts the sample IS (line $3$). Then, an inner loop is run for each network layer and for each layer channel to extract the MHAP of a training sample (lines $4$-$8$). The extracted MHAP is added to the cluster list (line $9$), and the cluster medians are returned as an output of the training phase (line $10$).

\subsection{MHAP Evoultion Graph}\label{Graph}
Inspired by the idea of the shapelet evolution graph~\cite{cheng2020time2graph}, we propose constructing the MHAP evolution graph. However, our MHAP evolution graph is different from the shapelet evolution graph both in nature and construction procedure. In MTS2Graph, we deal with more than one graph, each representing one CNN layer. After obtaining MHAPs from each network layer, time-aware graphs are constructed that maintain the occurrence order of MHAPs. As the evolving MHAP graphs represent each network layer, a merging strategy is required to create a graph representing the entire interpretable representatives. The steps to create a time-aware graph for each layer are as follows:

\subsubsection{Develop MHAP Graphs for each Layer:}
One layer MHAP time-aware graph is a directed and weighted graph $G$ consisting of nodes, $N$, representing the MHAP cluster medians of a layer ($CM_{MHAP}$) and edges, $E$, indicating the occurrence of an MHAP $n_i \in N$ followed by another MHAP $n_j \in N$ in a multivariate data. The weight of the edge $w_{i,j}$ indicates how often among different multivariate instances $n_i$ occurs followed by $n_j$.

To construct the MHAP evolution graph, we take the first MHAP of the input data and find the corresponding cluster from the $CM_{MHAP}$ list (e.g., cluster $a$). Then, the following MHAP is extracted and assigned to a cluster (e.g., $b$). These two samples represent two nodes in the graph ($n_a, n_b$) with a directed edge connecting $n_a$ to $n_b$ ($e_{a,b}$). This process is repeated for each multivariate sample in the training data. Hence, if an edge already exists in the graph, its weight increases by 1. Fig.~\ref{motivationImage}a shows sample data with the corresponding representative patterns of the first layer (MHAP). Fig.~\ref{motivationImage}c illustrates the generated first layer time-aware MHAP graph. This graph shows that MHAPs of the first sample are represented by node $29$, followed by node $25$, and last by node $10$, all connected via directed edges.

\begin{figure}[t]
\centering  
\includegraphics[scale=0.55]{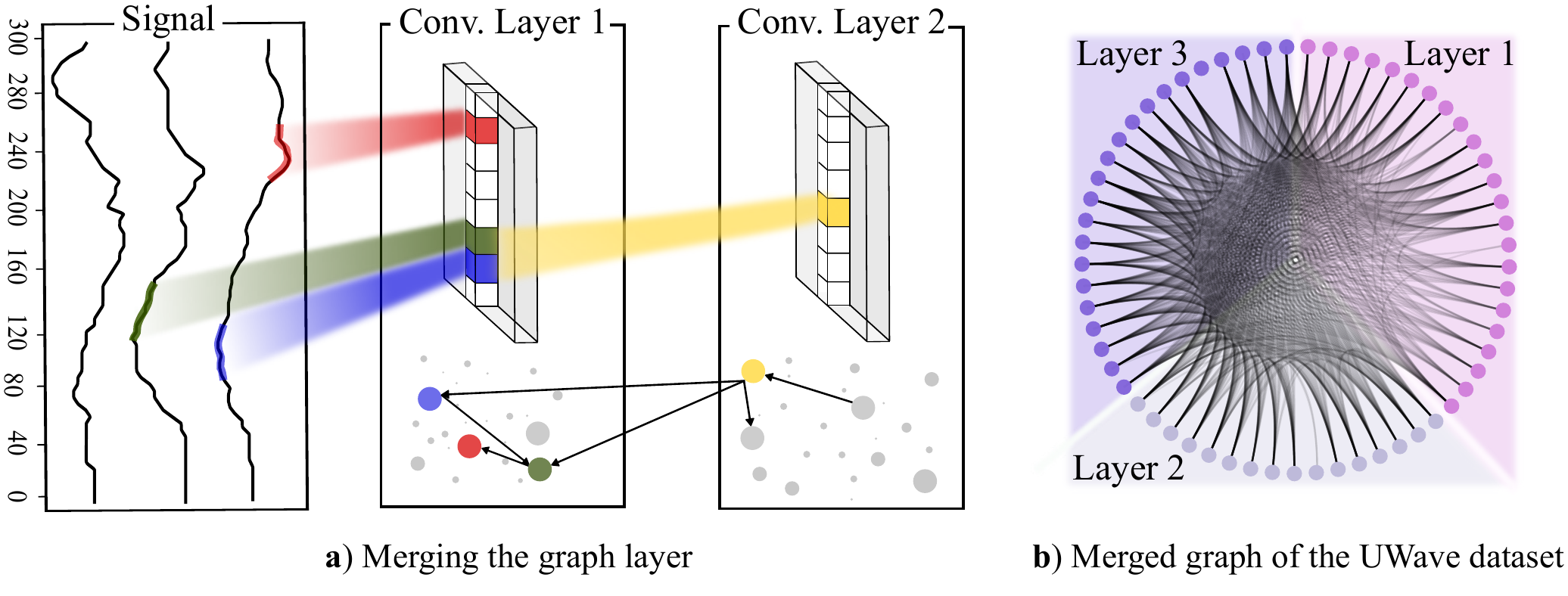}
%\includesvg{images/MotivationExample.svg}
\caption{a) A snapshot of how to merge the second time-aware graph with the first-layer graph, b) The final merged graph from a trained CNN network with three convolution layers on the UWave dataset.} \label{fig:LayersConnectionGraph}
\end{figure}

\subsubsection{Merge the Layers Evolution Graphs:}
Since each graph represents a network layer, a merging technique is required to merge these graphs into one. We propose a novel strategy to combine each layer's time-aware graph into a single one. This strategy preserves the correlation between the generated time-aware graphs of each layer while considering how the original CNN layers work. Fig.~\ref{fig:LayersConnectionGraph}a shows an example of two convolution layers and their corresponding time-aware evolution graphs. Each node in the graph represents an MHAP, and this MHAP refers to a collection of neurons from the previous layers, where the number of neurons depends on the assigned kernel size. For example, suppose the two convolutional layers in Fig.~\ref{fig:LayersConnectionGraph}a contain kernel sizes of $5$ and $3$, respectively. In that case, an MHAP in the first layer corresponds to a sequence of five input data points, and an MHAP in the second layer reflects three neurons from the first layer. Based on this, we merge each node in the graph with a node in the previous layer. For example, in Fig.~\ref{fig:LayersConnectionGraph}a, the yellow node in the second layer represents three neurons of the first layer, two of which are among the first-layer MHAPs, i.e., nodes in the first graph. Therefore, we connect the yellow node in the second-layer graph with the first-layer MHAPs (red and blue nodes). Fig.~\ref{fig:LayersConnectionGraph}b represents an example of a merged graph from a CNN with three layers for the UWave dataset. Note that Fig.~\ref{motivationImage}c is the first-layer graph of the same dataset before merging the layers' graphs.

\begin{algorithm}[tb] 
\caption{Generate MHAP Evolution Graph}
\label{alg:MHAPgraph}
\begin{algorithmic}[1]
\Require{$X_{train}$: Multivariate time series data} 
\Ensure{evolution graph $G$}
\State Model $\longleftarrow$ {CNN $(X_{train})$}
\For{$x$ \texttt{in $X_{train}$}}  
    \State $signal$ := InputSet($x$)

\For{$l$ \texttt{in Model\_layers}}
\For{$f$ \texttt{in $l$}} \Comment{filters in a CNN layer}
    %\State $MHAP_{layer}.add(get\_ MHAP(signal))$
    \State $HAN = [\forall n \in [1, N_{l,f}] \; where\; A(n) \geq T_{l,f}]$
    \State $MHAP_{layer}.add([Signal Receptive F ield(HAN)])$
    \EndFor
    \State $MHAP_{cnn}.add(MHAP_{layer})$
\EndFor
\EndFor
\State CM$_{list}:=K_{shape}(MHAP_{cnn}).Centroid$ 
\State $[c] := CM_{list}.size$ \Comment{vector of layer centroids size}
\For{l in \texttt{Model\_layers}}
    \State Initialize a layer graph $g$ with $c[l]$ nodes 
    \For{$x$ \texttt{in $X_{train}$}}  
        \For{all consecutive MHAPs in $x$} %\Comment{nodes $n_{i}$, $n_{i+1}$} %are $MCMHAP(MHAP)$}
             \State Add a directed edge $e_{t,t+1}$
        \EndFor
    \EndFor
    \State $G_{list}.add(g)$
\EndFor
\State $G$ := $merge(G_{list})$ 

\State \Return { $G$}
\end{algorithmic}
\end{algorithm}

Algorithm~\ref{alg:MHAPgraph} describes the process of creating a time-aware graph for each network layer and the merging strategy. The algorithm takes the training data as input and returns the time-aware evolution graph. First, a CNN-based model is trained using the training data (line $1$). A loop is executed for each training sample (lines $2$-$8$) and extracts the MHAP of a training sample. The extracted MHAP is then clustered, and the cluster centroids are added to a list (line $9$). To create a time-aware graph, first, for each CNN layer, a graph is initialized with the node size equal to the number of cluster centroids of that layer (line $12$). Then, a loop is executed (lines $13$-$15$) for each training sample, extracting MHAPs and assigning them to their respective cluster and connecting two adjacent MHAPs with a directed edge. Finally, the algorithm concatenated all the layers' graphs and merge them into one graph (line $17$). 

\subsection{Graph Embedding and Representation Learning}\label{learning}
After constructing the time-aware MHAP evolution graph, we apply a graph embedding algorithm to obtain node representations. First, we employ the DeepWalk algorithm \cite{perozzi2014deepwalk} to acquire the node representation vectors, $\Phi$ $\in$ $\mathbb{R} ^{D}$, where $D$ is the embedding size. 
%To learn these representations based on the new embedded graph, 
Next, we divide each multivariate sample into segments, and assign the corresponding MHAP to each segment. For each MHAP, $N_{i,j}$, we then retrieve its representation vector, $\Phi$($N_{i,j}$), and sum them over the segment. Then, we merge all these vectors into a representation vector $\bar\Phi_{x}$ for a time series sample. This new representation vector can be used as an input to any classifier. XGBoost \cite{chen2016xgboost} has shown promising results for univariate time series data~\cite{cheng2020time2graph}, hence, in MTS2Graph, We use it to classify the new multivariate time series representations. Algorithm~\ref{alg:graphebeddesudo} presents the detailed steps for Graph embedding and representation learning.
\begin{algorithm}[tb] 
\caption{Graph Embedding and Representation Learning}
\label{alg:graphebeddesudo}
\begin{algorithmic}[1]
\Require{$X_{train}$, $MHAP$, $G$} 
\Ensure{Graph embedding $\Phi$, MTS Representation Vector $\bar{\Phi}$}
\State $\Phi$:= Embeds graph $G$ with vector size $D$
\State $\bar{\Phi}$:= [] \Comment{Embedding vector of the training data}
\For{$x$ \texttt{in $X_{train}$}}
    \State Divide $x$ into $M$ segments
    \State  $\bar{\Phi}_{x}$ := [$0$]$_D$ 
    \For{$m$ \texttt{in $M$}}
        \State $\bar{\Phi}_{m}$ := [$0$]$_D$  
        \For{all $MHAP$ in $m$}
             \State $\bar{\Phi}_{m}$ += $\Phi$($MHAP$)
        \EndFor
        \State $\bar{\Phi}_{x}.add(\bar{\Phi}_{m})$
    \EndFor
    \State $\bar{\Phi}.add(\bar{\Phi}_{x})$
\EndFor

\State \Return {$\Phi$, $\bar{\Phi}$}

\end{algorithmic}
\end{algorithm}

\section{Results and Discussion}\label{results}
This section presents the result of our proposed MTS2Graph framework 
%\footnote{Link to our code \url{https://gitfront.io/r/user-8157177/Q1Csggct5jfW/MTS2Graph/}.} 
on eight time-series benchmarks from the UCR/UEA archive \cite{dau2019ucr,baydogan2019multivariate} %\cite{baydogan2019multivariate}
, along with HAR\cite{anguita2013public} and PAM 
\cite{reiss2012introducing} datasets. We run our experiments on a Core i5 8500 CPU PC with 32GB RAM.

\subsection{Experiments Setup}
%\textcolor{red}{this whole subsection is new}
%\input{images/table1}
We selected eight UCR/UEA, HAR, and PAM datasets to test the performance of our proposed MTS2Graph framework and compare it to the baselines and state-of-the-art methods. %Table~\ref{table0:archive} shows the statistics of each dataset. 
The UCR/UEA archive is a set of well-known benchmarks that contains both univariate and multivariate time series classification datasets. In this work, we choose a subset of this archive set with a various number of samples, dimensions (channels) and classes. The selected datasets contain a larger number of samples and lengthier time series among the 13 datasets from the archive. Those datasets with less than 100 samples or time series lengths were excluded since they are not much interesting for investigations. 
The Human Activity Recognition (HAR) dataset records the daily activities of 30 participants and generates six different labels for these activities. %(lying down, sitting, standing, walking, walking upstairs, and walking downstairs). 
The Physical Activity Monitoring (PAM) dataset recorded 18 different daily activities with 17 sensors. In this study, we used only 8 activities with more than 500 samples. 
 
For all datasets, we run 10-fold cross-
validation with $80\%$ for training, 10$\%$ for validation, and 10$\%$ for test data. Each experiment is run $500$ epochs, with $100$ embedding size, and a segment length of size $10$. We compare MTS2Graph's predictive performance to several benchmark approaches and use their proposed network architecture and parameters to reproduce similar results%\footnote{The additional materials include more details on the datasets and the baseline methods}. 
\begin{table}[tb]
%\begin{table*}
\caption{Comparison of classification performance on the selected datasets. The best accuracy among all models is highlighted.}
%\centering
%\renewcommand{\arraystretch}{2.1}
\resizebox{\textwidth}{!}{%
\begin{tabular}[ht]{lccccccccccccc}
 \toprule
& \multicolumn{11}{c}{ Public Datasets }  &  \multicolumn{1}{c}{}  &   \\
\cmidrule{2-11}\cmidrule{13-14}
%\Bstrut
 Model & UWave  & ArabicDigits & AUSLAN  & Libras  & ECG  & NetFlow & Wafer & CharacterTrajectories&HAR&PAM& & Avg.Rank& Wins/Ties \\  
  \hline
  
%\Tstrut
 1NN-ED& 0.881 & 0.967 & 0.700  &0.833   & 0.900 & 0.798  &0.975 &0.964 &0.983 &0.275 && 10.6&0\\
 
1NN-ED (Norm)&0.881  & 0.967 & 0.677 & 0.833  & 0.850  &0.873 & 0.908&0.964  &0.990&0.568 &&11.1& 0 \\

1NN-DTW-I& 0.869 & 0.960 & 0.747 & 0.894 &0.850 &0.791  &0.983  &0.969 &0.994 &0.919 &   &8.7& 0 \\

1NN-DTW-I (Norm)&  0.868 &0.959  &0.894 & 0.833  & 0.75 &0.895 & 0.966& 0.969 &\textbf{0.999} &0.859&  &10.0&1 \\

1NN-DTW-D& 0.903&0.963 &0.750& 0.872 & 0.800 & 0.753  & 0.975 &0.990 &\textbf{0.999} &0.476&&9.0& 1  \\

1NN-DTW-D (Norm)&0.903 &0.963 &0.731 &0.870  & 0.800 &0.925  &0.933  &0.989 &0.995 &0.699 & &9.2&0  
%\Bstrut
\\\hline
%\Tstrut
WEASEL+MUSE\cite{schafer2017multivariate}&0.916 & \textbf{0.990}& 0.970& 0.894 &  0.880 & 0.961 &  \textbf{0.997} &0.990 &N/A&N/A&  &5.6&2   \\

GMSM$_{red}$\cite{medico2021learning}& 0.945 & 0.989&\textbf{0.988} &0.863 & 0.750 &  0.805 & 0.908& 0.996 &0.959 &0.833 & &8.1&1\\

ShapeNet\cite{li2021shapenet}&0.906& 0.980 & 0.950 & 0.856  & 0.620 & 0.970 &0.882  &0.980&N/A&N/A&  &10.5& 0 
%\Bstrut
\\\hline
%\Tstrut

FCN\cite{ismail2019deep}& 0.930 & \textbf{0.990}  & 0.970 & 0.960 & 0.870 & 0.890 & 0.980 &0.993 &0.976&\textbf{0.958} & &4.0&2  \\

LSTM\cite{6795963}& 0.948  & 0.981 & 0.424 & 0.720 & 0.850 & 0.880& 0.925 &0.979 &0.965 &0.713&  &9.7& 0 \\

TapNet\cite{zhang2020tapnet}& 0.891  &  0.865& 0.897 & 0.883 & 0.740& 0.822 & 0.957 &\textbf{0.997}&0.961&0.857&&9.9&1   \\

MLSTM-FCN\cite{karim2019multivariate}&\textbf{0.970} & \textbf{0.990}& 0.950 & 0.970 & 0.870 &  0.950& 0.990 &0.985 &0.967&0.900&  & 4.3& 2    \\ 

$C^2$AF\cite{bai2021correlative}& 0.899 &  0.975 & 0.103 & 0.708 & 0.825 & 0.955& 0.97& 0.826 &0.586&0.857 & &11.1&  0  \\ 

GTN\cite{liu2021gated}& 0.910 & 0.980 & 0.970 & 0.889 & 0.910  &  \textbf{1.00}& 0.991  & 0.970 &0.445 &0.353 && 6.5 &  1  
%\Bstrut
\\\hline
%\Tstrut
\textbf{MTS2Graph}& 0.930 &\textbf{0.990}  &0.950 & \textbf{0.979} &\textbf{0.93}  &\textbf{1.00} & 0.976  &0.990 &0.974 &0.860 & &\textbf{3.5}&\textbf{4}   \\
\bottomrule

\end{tabular}%
}
\label{table0:accuercy}
\end{table}

\subsection{Predictive Performance Comparison Results}
\begin{figure}[t]
\centering
\includegraphics[scale=0.60]{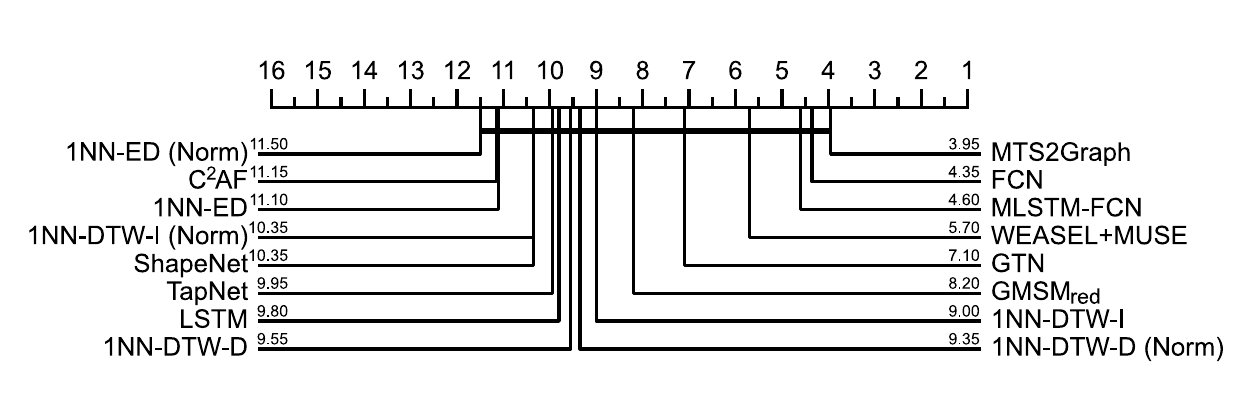}
\caption{Critical difference diagram for the used multivariate time series classifiers on ten MTS data.}
\label{cd-digram}
\end{figure}

We compare the predictive accuracy of MTS2Graph to several comparison methods on various benchmark datasets in Table~\ref{table0:accuercy} and report the number of wins/ties and their average rank%\footnote{Detail on how we compute the average rank is in the additional material.} 
for each approach. Note that the N/A in the table indicates  the experiments where the model runs out of memory. 
The wins/ties indicator shows that our proposed MTS2Graph framework outperforms the comparison methods on the ECG and Libras datasets and shares the first rank (and best performance) with some other comparison methods in two other datasets (ArabicDigits and NetFlow). On UWave, AUSLAN, Wafer, CharacterTrajectories, and PAM datasets, MTS2Graph performs slightly worse than the best-reported accuracy. It performs better than distance-based models and at least three deep-learning-based methods. For the HAR dataset, our method performs slightly worse than the best accuracy reported from the two distance-based models, 1NN-DTW-I (Norm) and 1NN-DTW-D. %From its average rank indicator,
We observe that our proposed method, with a 3.5 average rank, outperforms all the other baseline and state-of-the-art methods. The best state-of-the-art approaches, FCN, MLSTM-FCN, and WEASEL+MUSE, achieve an average rank of $4.0$, $4.3$, and $5.6$, respectively. 
Overall our proposed MTS2Graph framework shows satisfactory performance on the tested benchmarks, both with small and large dimensions and various sample sizes. Moreover, it benefits from network decision interpretability over the methods of comparison. Figure~\ref{cd-digram} shows the critical difference diagram, which illustrates the average rank of the classification algorithms tested on different datasets. The solid bar represents the cliques and indicates no significant difference between the classifiers within the same clique. Comparison of classifier performance is based on pairwise Wilcoxon signed rank tests, and cliques are determined using the Holm correction. Since MTS2Graph is developed as an interpretable classifier for MTS data, lower accuracy is expected. However, the diagram shows that MTS2Graph performs comparably to other classifiers and achieves the highest accuracy among all tested classifiers.

\begin{figure}[t]
\centering
\includegraphics[scale=0.55]{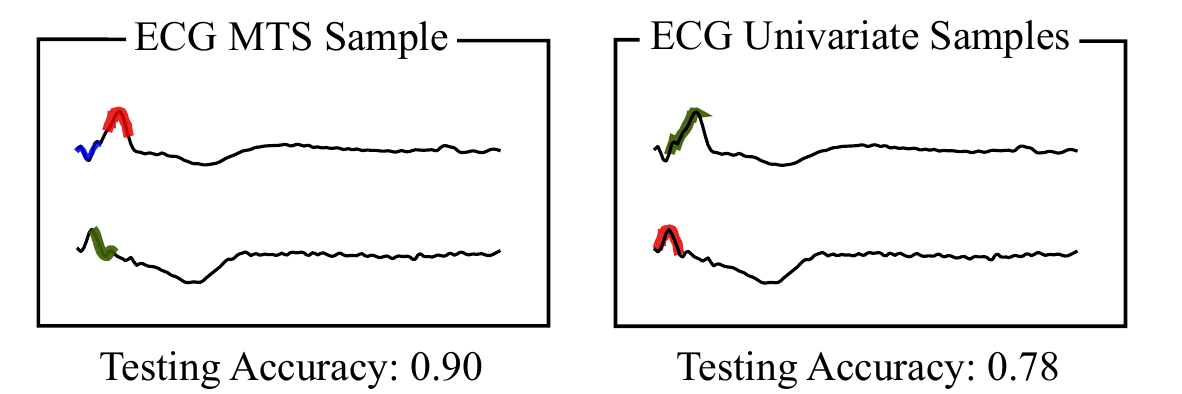}
\caption{A two-variate ECG data example showing the difference between the extracted MHAPs when taking the data as MTS and taking each variate independently.}
\label{casestudy}
\end{figure}

\begin{figure*}[t]
\includegraphics[scale=0.55]{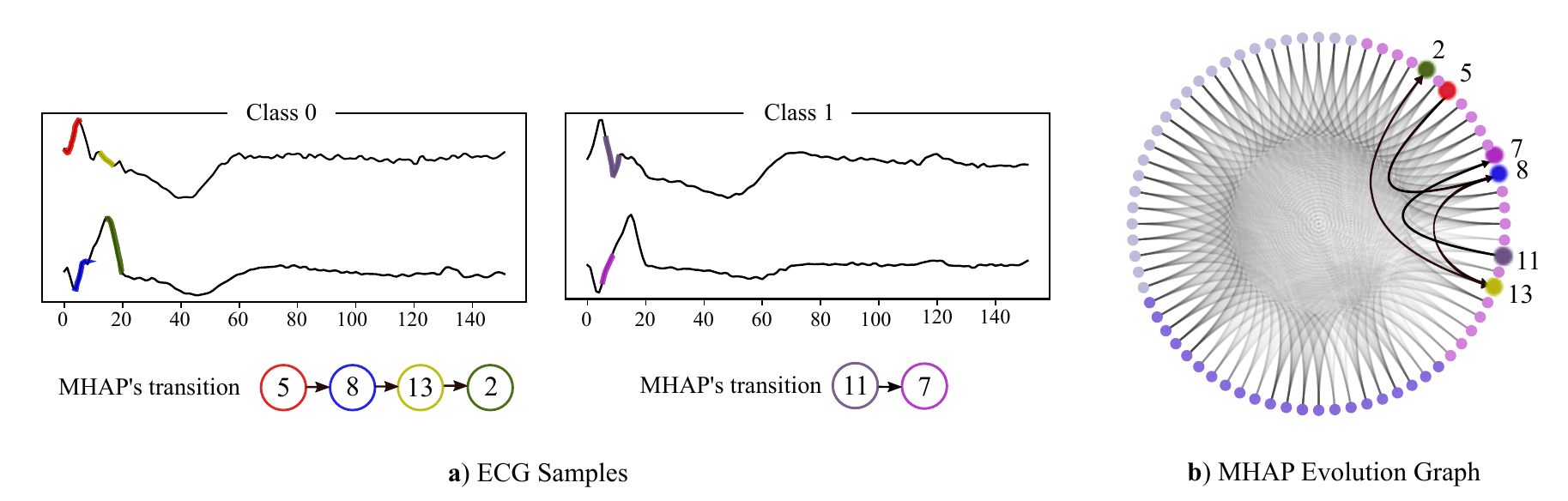}
\caption{a) Two random instances from the two-variate ECG dataset and their highly activated subsequences from the first CNN layer, b) the MHAP evolution graph that is built based on the timely occurrence order of the extracted MHAPs.}
\label{exampleecg}
\end{figure*}
%Two random instances from the two-variate ECG dataset and their highly activated subsequences from the first CNN layer,a) sample with an output class label $0$, andt class label $1$, b) the MHAP evolution graph that is built based on the timely occurrence order of the extracted MHAPs. The colored nodes represent the respective MHAPs of the two instances in a).
We conducted a time complexity analysis of MTS2Graph on the tested datasets. Our findings reveal that the most time-consuming task is training a convolutional neural network (CNN), which accounts for 48.1\% of the total time. On the other hand, creating the extended input set, takes the least time, comprising 2.3\% of the total time, as it does not require additional training on the extended set. Instead, it predicts the network accuracy for all possible signal combinations. Extracting MHAPs from the data takes up approximately 20.4\% of the total time. This task's time complexity depends on the data's size, with larger data sets requiring more time to find input data periods that highly activate the neurons. This outcome is anticipated since the search space becomes broader and demands more time. The remaining time is distributed among other components of the MTS2Graph framework, with approximately 23\% used for training the XGboost classifier and 6.4\% for clustering MHAPs.

\subsection{Medical Case Study: ECG Data}
In this section, we present a case study of the ECG dataset to investigate the interpretability and effectiveness of the MTS2Graph framework. Fig.~\ref{casestudy} shows an example of the difference in the model's performance when the test data is fed to the model as a multivariate time series (MTS) sample or multiple univariate samples. Fig.~\ref{casestudy} left shows the extracted MHAPs when the correlation among the input variants is taken into account, and the MTS data is taken as a whole in the model training process. We can observe that this data sample has three representative patterns, and the model accuracy achieves $0.90$. However, in Fig.~\ref{casestudy} right, we feed the MTS input as two independent univariate samples and ignore the correlation between the signals. As a result, the extracted MHAPs differ from the previous experiment, and the model accuracy is also decreased to $0.78$. This example indicates the importance of using the multivariate time series data and considering its variable dependencies for the classification tasks.

Fig.~\ref{exampleecg} shows an example of the MHAP evolution graph for the ECG data and the MHAP transitions in time for two samples from classes 0 and 1. We observe that the MTS2Graph framework not only outputs the important subsequences of the multivariate time series data but also represents the temporal transition using the MHAP evolution graph. Fig.~\ref{exampleecg}a presents the four extracted MHAPs for the class 0 sample and two extracted MHAPs for the class 1 sample. These MHAPs are also nodes in the graph in Fig.~\ref{exampleecg}b. Node transitions show the temporal order of MHAPs' occurrence in the input data. MTS2Graph relies on the important subsequences (MHAPs) of the input data and considers their temporal order, as this is essential for understanding the network's decision in classifying time series. For example, in ECG data reading, the physician must read the data with a specific sequence order in time to decide whether the patient's ECG record is normal or has some pathology issues. %The second ECG example in Fig.~\ref{exampleecg}a) belongs to class 1 and also shows the extracted MHAPs and their temporal transitions in the graph. Unlike the first example, this data example contains only two nodes that can help the network identify the output class.

%\begin{figure}[t]
%\centering
%\includegraphics[scale=0.6]{images/embadECG.pdf}
%\caption{Example of MHAP evolution graph Node %Embedding for ECG data.}
%\label{nodeembadding}
%\end{figure}
%Fig.~\ref{nodeembadding} shows the ECG MHAP evolution graph embedding of Fig.~\ref{exampleecg}b in the 2D space where these embedding show that each layer embedding belongs to the same cluster. The new embedding vectors define the node representatives used as input to XGBoost.

\subsection{Parameter Sensitivity Analysis}
\begin{figure}[t]
\centering
\includegraphics[scale=0.6]{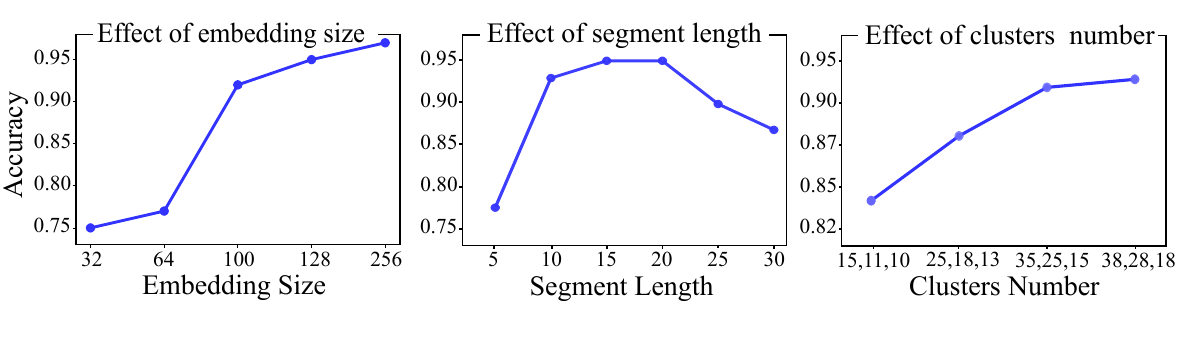}
\caption{Parameter sensitivity analysis for ECG dataset.}
\label{sensitivity}
\end{figure}

MTS2Graph framework has three primary hyperparameters: the segment length, the graph embedding size and the clusters number. In this section, we investigate the sensitivity of our MTS2Graph framework to these hyperparameters and report the results for the ECG dataset (Fig.~\ref{sensitivity}). Experiment results for the embedding vector size show that a smaller vector ($32$, $64$) has a less representative quality and reduces the predictive accuracy of MTS2Graph by about $0.15$. Accuracy is improved for a larger embedding size; however, a high dimensional feature vector can increase the complexity of XGBoost or any other classifier. Therefore, selecting an appropriate size for the graph embedding is crucial to balance the trade-off. Fig.~\ref{sensitivity} shows the segment length's effect and indicates that the best segment length for the ECG dataset should be selected between $10$ and $20$, which is about $10\%$ of the input length. This is because much smaller segment lengths may discard the useful MHAPs, and the larger segments could extract overfitted MHAPs. Figure~\ref{sensitivity} depicts our investigations on cluster size. Since we use a three-layer CNN, there are three parameters in this regard. The figure shows the effect of these parameters and that the best cluster size selection is ($l_1 = 38$, $l_2 = 28$, $l_3 = 18$). %Due to the reduced number of MHAPs in the upper layers of CNN, it is intuitive that the cluster size in those layers should be set less than the lower layers. 
We chose these numbers by trial and error. For any new dataset, one can use a validation set and apply the elbow method \cite{ng2012clustering} to find the best cluster numbers. 

\section{Conclusion} \label{conclusion}
In this paper, we presented MTS2Graph, a novel framework for interpreting the multivariate time series classification problem by extracting the multivariate highly activated input sequence (MHAP) of a CNN model and constructing a time-aware graph. In order to achieve our goal, we extract the input subsequences that fire the neurons of a CNN network channel, then build a graph to obtain the temporal correlation between the extracted MHAPs, and use a graph embedding algorithm to learn the new feature representation. %As a result, our framework can classify multivariate time series and identify the most important features for a classifier. 
Our experiments on several datasets indicate the effectiveness of MTS2Graph. In the future, we will improve our method to perform in evolving manner by maintaining the MHAPs and classifying them upon the occurrence of new patterns or changes in the existing patterns.
%we will extend the explainability of our framework by identifying the extracted MHAPs' semantic annotations using domain knowledge experts and testing the framework on more complex real-case applications. Also, we will improve our method to perform in evolving manner by maintaining the MHAPs and classifying them upon the occurrence of new patterns or changes in the existing patterns.

%\subsubsection{Acknowledgements} This work was supported by German Federal Ministry of Education and Research (BMBF) under grant agreement No. 01IS19063A (project HAISEM-Lab, 2019-2022).

\bibliographystyle{splncs04}
\bibliography{bibdata}

\end{document}